\documentclass{article}
\usepackage{spconf,amsmath,epsfig}

\usepackage{multirow}  
\usepackage{makecell}  
\usepackage{times}
\usepackage{epsfig}
\usepackage{graphicx}
\usepackage{amssymb}
\usepackage{multirow}
\usepackage{enumitem}
\usepackage{xspace}
\usepackage{cite}
\usepackage{url}
\usepackage{enumerate}
\usepackage{color}

\makeatletter

\DeclareRobustCommand\onedot{\futurelet\@let@token\@onedot}
\def\@onedot{\ifx\@let@token.\else.\null\fi\xspace}

\def\etal{\emph{et al}\onedot}

\begin{document}

\title{Single image depth estimation by dilated deep residual convolutional neural network and soft-weight-sum inference}

\name{Bo Li$^{1*}$, Yuchao Dai$^{2}$, Huahui Chen$^{1}$,  Mingyi He$^{1*}$
      \thanks{$^{*}$ libo.npu@gmail.com}
\address{$^{1}$School of Electronics and Information, Northwestern Polytechnical University, China \\
         $^{2}$Research School of Engineering, Australian National University, Australia \\
         }
}

\maketitle
\begin{abstract}
{This paper proposes a new residual convolutional neural network (CNN) architecture for single image depth estimation. Compared with existing deep CNN based methods, our method achieves much better results with fewer training examples and model parameters. The advantages of our method come from the usage of dilated convolution, skip connection architecture and soft-weight-sum inference. Experimental evaluation on the NYU Depth V2 dataset shows that our method outperforms other state-of-the-art methods by a margin.}
\end{abstract}

\section{Introduction}
Single image depth estimation aims at predicting pixel-wise depth for a single color image, which has drawn increasing attentions in computer vision, virtual reality and robotic. It is a very challenging problem due to its ill-posedness nature. Recently, there have been considerable efforts in applying deep convolutional neural network (CNN) to this problem and excellent performances have been achieved\cite{li2015depth,liu2015deep,wang2015towards,eigen2015predicting,cao2016estimating}.

Li \etal~\cite{li2015depth} predicted the depth and surface normals from a color image by regression on deep CNN features in a patch-based framework. Liu \etal~\cite{liu2015deep} proposed a CRF-CNN combined learning framework. Wang \etal~\cite{wang2015towards} proposed a CNN architecture for joint semantic labeling and depth prediction. Recent works have shown that the depth estimation problem could benefit from a better CNN architecture design. Eigen \etal~\cite{eigen2015predicting} proposed a multi-scale architecture that first predicts a coarse global output and then refines it using finer-scale local networks. Very recently, Cao \etal~\cite{cao2016estimating} demonstrated that formulating depth estimation as a classification task is better than direct regression.
These works demonstrate that, network architecture design plays a central role in improving the performance. 

To this end, a simple yet effective dilated deep residual CNN architecture is proposed, which could converge with much fewer training examples and model parametres. Furthermore, we analyze the statistic property of our network output, and propose soft-weight-sum inference instead of the traditional hard-threshold method. At last, we conduct evaluations on widly used NYU Depth V2 dataset~\cite{Silberman2012Indoor}, and outperforms other state-of-the-art methods by a margin.
 
\section{Network Architecture}
\begin{figure}[ht]
\centering{\includegraphics[width=0.7\linewidth]{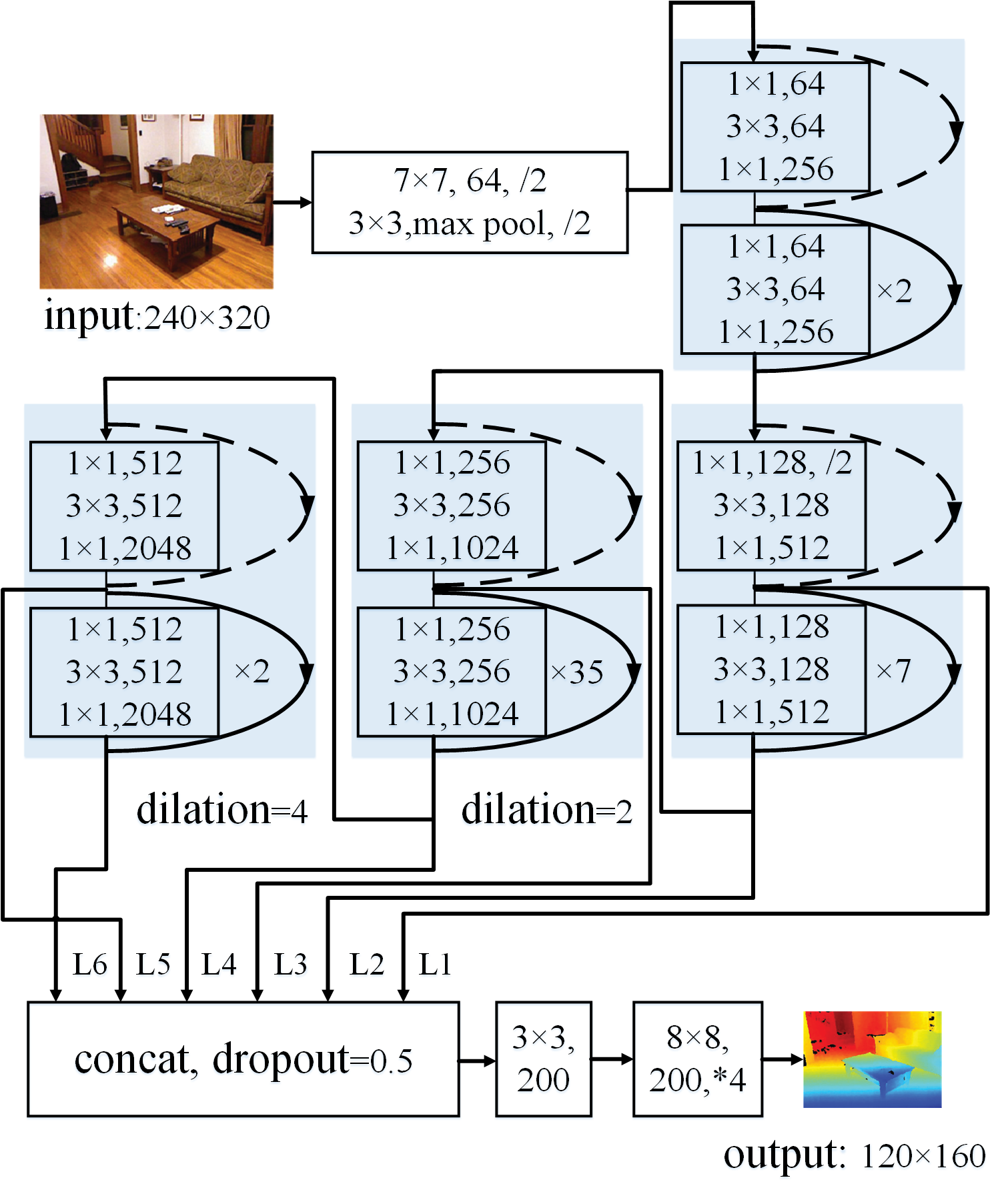}}
\caption{Illustration of our network architecture. The detail of the basic residual block could be refer to~\cite{ResNet}. $\times n$ means the block repeats $n$ times. We present all the hyper-parameters of convolution and pooling layers. All the convolution layers are followed by batch normal layer except the last one.  $/2$ means the layer's stride is 2. $*4$ means the deconv layer's stride is 4. Dilation shows the dilated ratio of the correspondent parts. $L1, \cdots L6$ are our skip connection layers.}
\label{network}
\end{figure}

Our CNN architecture is illustrated in Fig. \ref{network}, in which the weights are initialized from a pre-trained 152 layers residual CNN\cite{ResNet}. The original network~\cite{ResNet} was specially designed for image classification problem. In this work, we transform it to be suitable to our depth estimation task. 

Firstly, we remove all the fully connect layers. In this way, we greatly reduce the number of model parameters as more than 80\% of the parameters are in the fully connect layers\cite{eigen2015predicting,cao2016estimating}. Although, both \cite{eigen2015predicting} and \cite{cao2016estimating} preserved the fully connect layers for long range context information, our experiments show that it is unnecessary in our network for the usage of dilated convolution. Secondly, we take advantage of the dilated convolution, which could expand the receptive field of the neuron without increasing the parameters. Thirdly, with dilated convolution, we could keep the spatial resolution of feature maps. Then, we concatenate intermediate feature maps with the final feature map directly. This skip connection design benefits the multi-scale feature fusion and boundary preserving.

(1) Dilated Convolution: Recently, dilated convolution is successfully utilized in the CNN design by Yu \etal\cite{yu2015multi}. Specially:

Let $F : \mathcal{Z}^2 \rightarrow \mathcal{R}$ be a discrete function. Let $\Omega_r = \left[ r,r\right] ^2 \cap \mathcal{Z} ^2 $ and let $ k: \Omega_r \rightarrow \mathcal{R}$ be a discrete filter of size $ \left( 2r+1 \right)^2  $. The discrete convolution filter $*$ can be expressed as
\begin{align}
\label{convolution}
( F * k ) (\mathbf{p}) = \sum\limits_{\mathbf{s} + \mathbf{t} = \mathbf{p}}F(\mathbf{s})k(\mathbf{t}).
\end{align}
We now generalize this operator. Let $l$ be a dilation factor and let $\mathbf{*}_l$ be defined as
\begin{align}
\label{dilated convolution}
( F *_l k ) (\mathbf{p}) = \sum\limits_{\mathbf{s} + l\mathbf{t} = \mathbf{p}}F(\mathbf{s})k(\mathbf{t}).
\end{align}

We refer to $ \mathbf{*}_l $ as a dilated convolution or an $l$-dilated convolution. The conventional discrete convolution $*$ is simply the 1-dilated convolution. 
%
%

(2) Skip Connection: As the CNN is of hierarchical structure, which means high level neurons have larger receptive field and more abstract features, while the low level neurons have smaller receptive field and more boundary information. We propose to concatenate the high-level feature map and the inter-mediate feature map. The skip connection structure benefits both the multi-scale fusion and boundary preserving.
 


\section{Soft-Weight-Sum Inference}
We reformulate depth estimation as classification task by equally discretizing the depth value in log space as~\cite{cao2016estimating}. Specially, we train our network with the multinomial logistic loss $E = -\frac{1}{N} \sum_{n=1}^N {\log(p_{k}^n)} $ followed by softmax $p_{i} = \frac {\exp{x_{i}}} {\sum_{i^{'}=1}^{m}{\exp{x_{i^{'}}}}}$. Here, $N$ is the number of training samples, $k$ the correspondent label of sample $n$, $m$ is the number of bins.



\begin{figure}[htb]
\centering
\scalebox{1}{
\begin{tabular}{@{}c@{}c}

\includegraphics[width=0.45\linewidth]{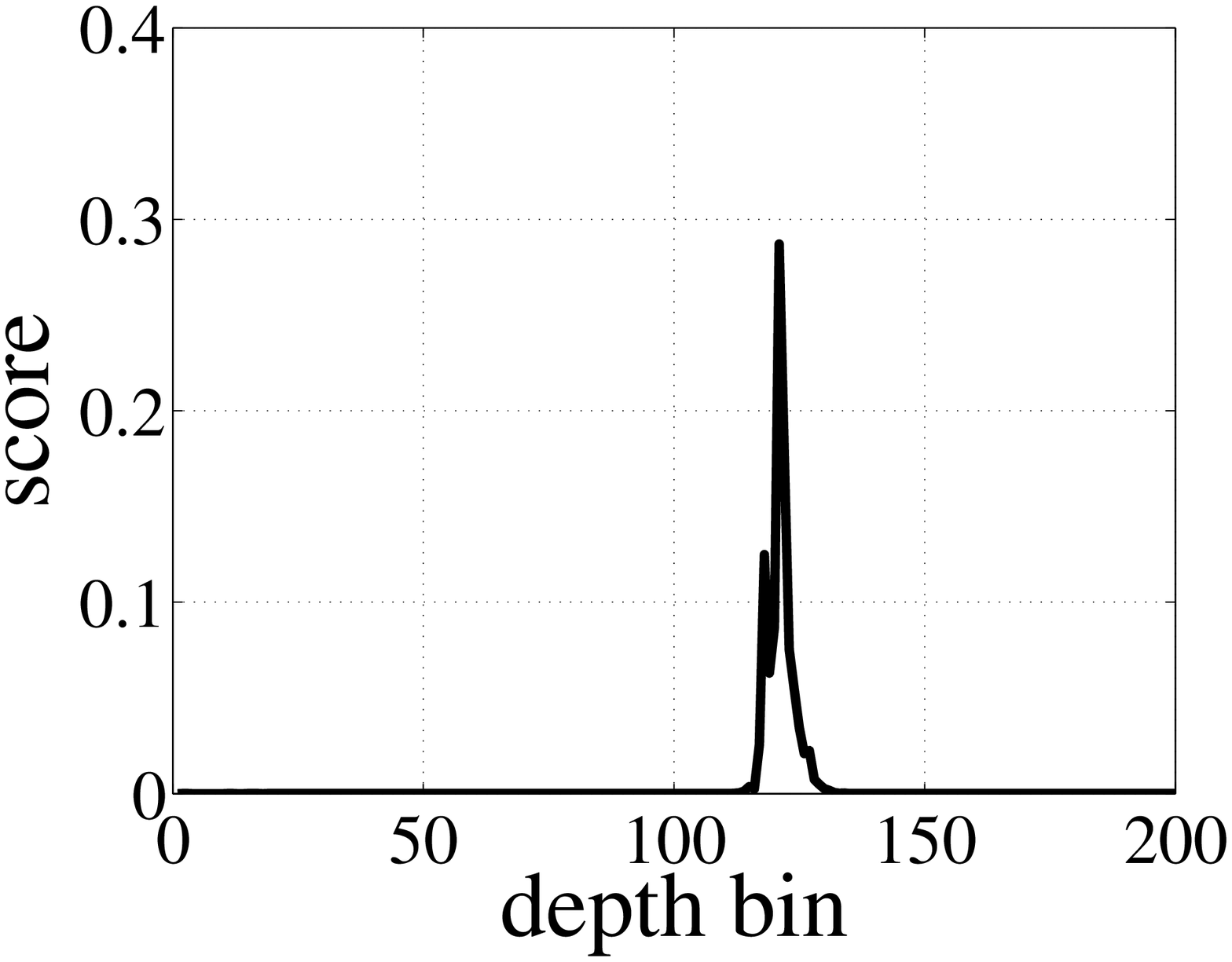} & \includegraphics[width=0.43\linewidth]{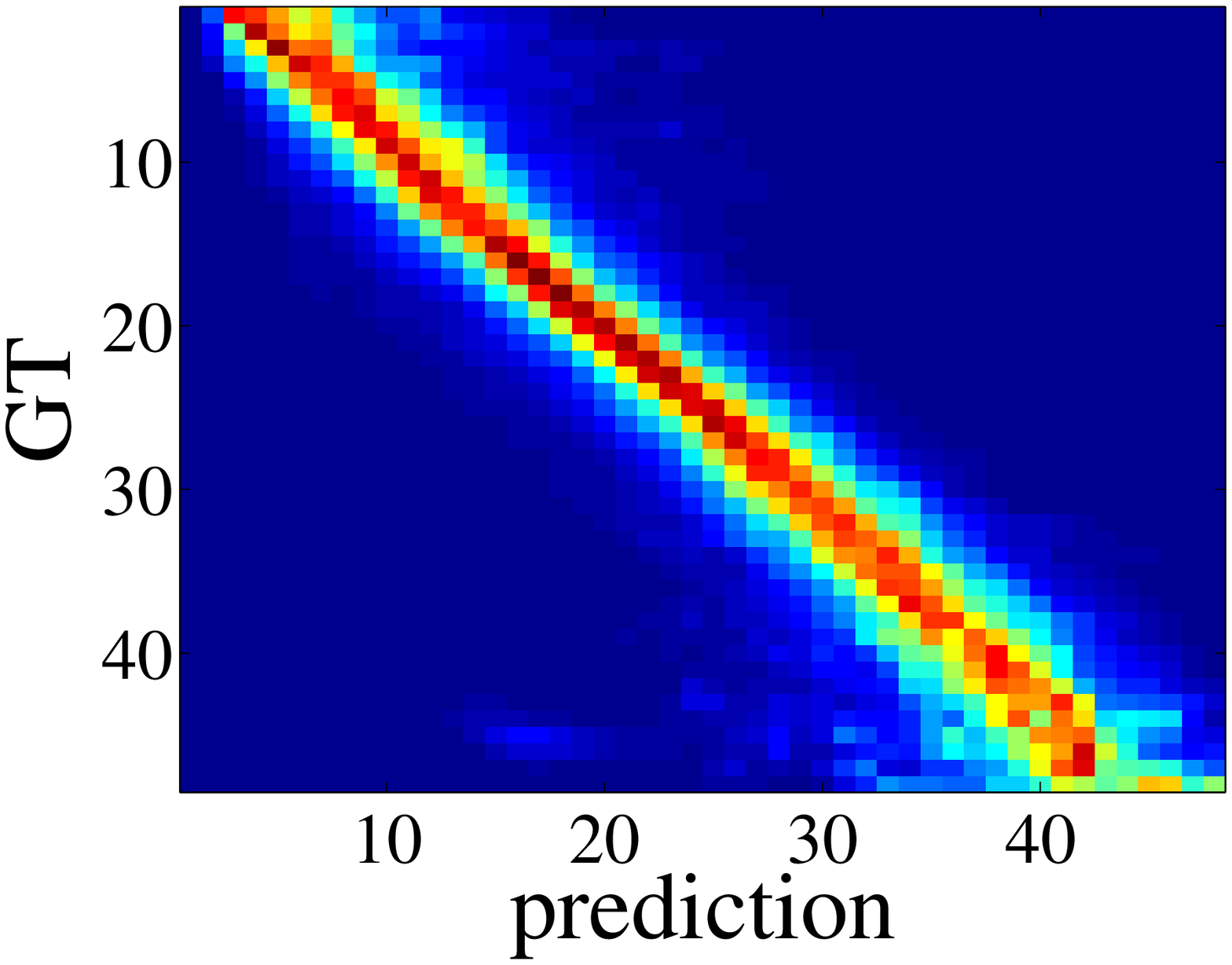}\\
a & b\\
\end{tabular}
}
\caption{Illustration of our model prediction. 
a Typical score distribution of network output.
b Confusion matrix on the test set, where we transfer the 200 bins to 50 for better illustration.}
\label{fig:score}
\end{figure}

A typical predicted score distribution is given in Fig. \ref{fig:score}a, where the non-zero score is centralized. Fig. \ref{fig:score}b is the confusion matrix on the test set, which present a kind of diagonal dominant structure. In Table \ref{tab: num bins}, we give the pixel-wise accuracy and Relative error (Rel) with respect to different number of discretization bins.

These statistic results show that: Even though the model can't distinguish the detailed depth well, it still learns the correct  concept of depth as the non-zero predicted score is centralized and around the right label. These statistic results also explain why increasing the number of bins could not improve the performance further, mainly due to the decrease of the pixel-wise accuracy. 

\begin{table}[htb]
\Huge
\caption{Pixel-wise accuracy and Rel w.r.t. number of bins.}
\resizebox{.65\linewidth}{!}
{
\begin{tabular}{ | l | c | c | c | c | c |}
\hline
num of bins              &50 &100 &200 &500 &1000\\
\hline
pixel accuracy (\%)           &67   &41  &25 &12  &7 \\
Rel                      &0.179   &0.137  &0.125  &0.127  &0.126 \\
\hline
\end{tabular}
}
\label{tab: num bins}
\end{table}

Inspired by these statistic results, we propose the soft-weight-sum inference. It is worth noting that, this method transforms the predicted score to the continuous depth value in a nature way. Specially:
\begin{equation}
\label{soft-weight-sum}
d_i = \exp{ \{ \mathbf{w}^{T} \mathbf{p}_i \}},
\end{equation}
where $ \mathbf{w} $ is the weight vector of depth bins. $\mathbf{p}_i $ is the output score for sample $i$. In our experiments, we set the number of bins to 200.

\section{Experiments}
We test our method on the widely used NYU V2 dataset~\cite{Silberman2012Indoor}. The raw dataset consists of 464 scenes, captured with a Microsoft Kinect, The official split consists of 249 training and 215 test scenes. 
We equally sample frames out of each training sequence, resulting in approximately 12k unique images. After off-line augmentations, our dataset comprises approximately 48k RGB-D image pairs.

\textbf{Implementation details:} Our implementation is based on the CNN toolbox: caffe~\cite{jia2014caffe} with an NVIDIA Titian X GPU. The proposed network is trained by using stochastic gradient decent with batch size of 3 (This size is too small, thus we average the gradient of 5 iterations for one back-propagation), momentum of 0.9, and weight decay of 0.0004. Weights are initialized by the pre-trained model from ~\cite{ResNet}. The network is trained with iterations of 50k by a fixed learning rate 0.001 in the first 30k iterations, then divided by 10 every 10k iterations.

For quantitative evaluation, we report errors obtained with the following error metrics, which have been extensively used. Denote $d$ as the ground truth depth, $\hat{d}$ as the estimated depth, and $T$ denotes the set of all points in the images.

\begin{itemize}[noitemsep,nolistsep]
  \item Threshold: $\%$ of $d_i$ {\rm s.t.} $ \max\left(\frac{\hat{d}_i}{d_i},\frac{d_i}{\hat{d}_i}\right)=\delta<thr$;
\item Mean relative error (rel): $\frac{1}{\vert T \vert}\sum_{d \in T}\vert{\hat{d}-d}\vert/d$;
\item Mean ${\log}_{10}$ error ($\mbox{log}_{10}$):\\
  $\frac{1}{\vert T \vert}\sum_{d \in T}\vert{\log_{10}\hat{d}-\log_{10}d}\vert$;
\item Root mean squared error (rms):\\
  $\sqrt{\frac{1}{\vert T \vert}\sum_{d \in T}{\Vert{\hat{d}-d}\Vert}^2}$.
\end{itemize}
Here $d$ is the ground truth depth, $\hat{d}$ is the estimated depth, and $T$ denotes the set of all points in the images.

\begin{figure}[!htb]
\centering
\scalebox{0.6}{
\begin{tabular}{@{}c@{}c@{}c}

\includegraphics[width=0.5\linewidth]{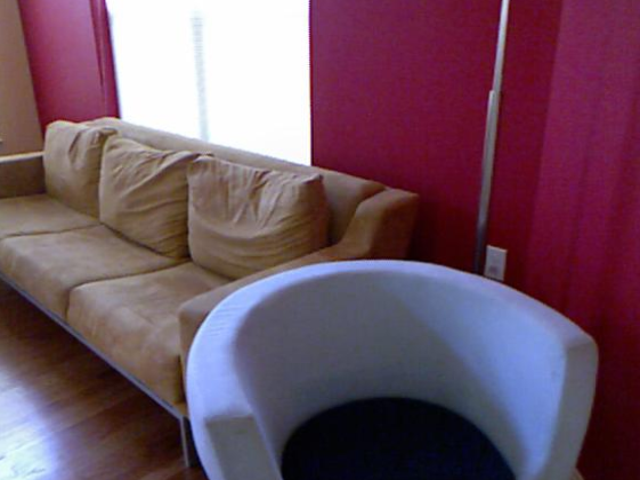}\ \ \ \ \
& \includegraphics[width=0.5\linewidth]{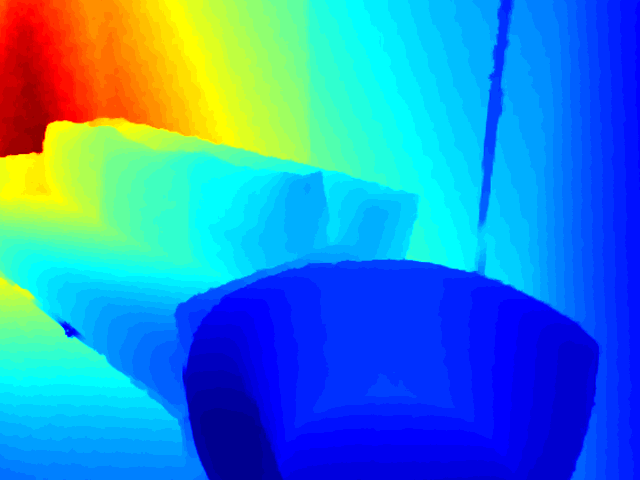}\ \ \ \ \
&\includegraphics[width=0.5\linewidth]{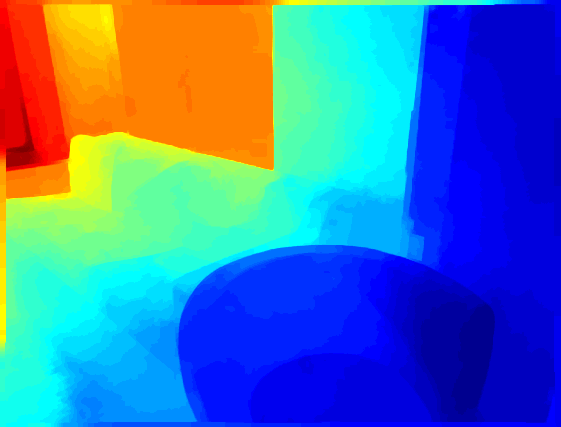}\\

\large{RGB} &\large{GT} &\large{Li \etal~\cite{li2015depth}}\\

\includegraphics[width=0.5\linewidth]{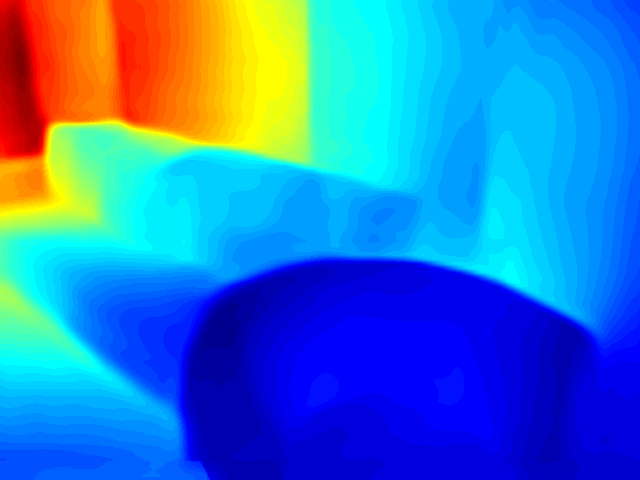}\ \ \ \ \
&\includegraphics[width=0.5\linewidth]{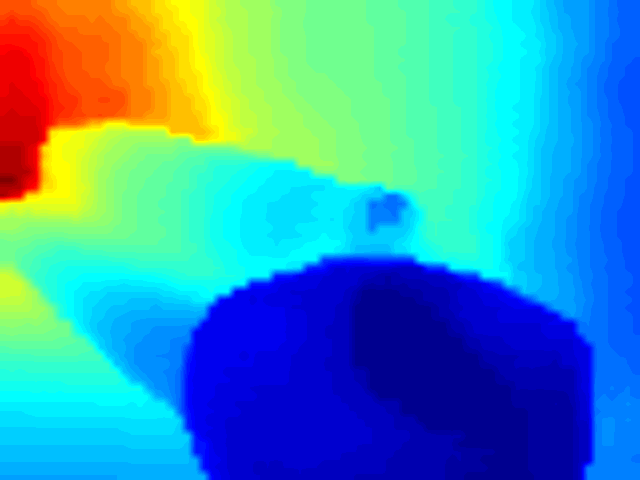}\ \ \ \ \
& \includegraphics[width=0.5\linewidth]{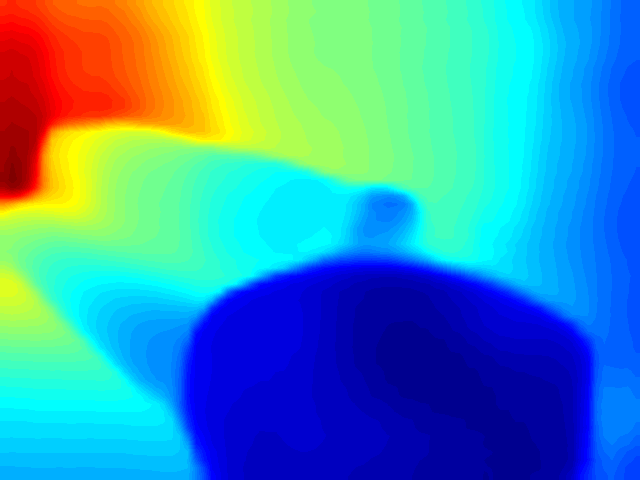} \\

\large{Eigen \etal~\cite{eigen2015predicting}} &\large{Ours hard} &\large{Ours soft}\\

\end{tabular}
}
\caption{Qualitative comparison of the estimated depth map on the NYU V2 dataset with our method and state-of-the-art methods. Color indicates depth (red is far, blue is close).}
\label{fig:nyu2}
\end{figure}

Quantitative and qualitative evaluation of our method is presented in Table \ref{tab:nyu2} and Fig. \ref{fig:nyu2} respectively. Our method outperforms other state-of-the-art depth estimation methods by a large margin with much fewer training examples and model scale. It is worth noting that, without any post processing, our result is of high visual quality. 

To further evaluate our method, we conduct experiments to analyze the contribution of each component and the results are illustrated in Table \ref{tab:analysis}. From Table \ref{tab:analysis} we can see that dilated convolution, skip connection, and soft-weight-sum inference all contribute to final depth estimation. From Fig. \ref{fig:nyu2}, we could also observe that the soft-weight-sum inference is beneficial to smooth the depth map while keeping the boundary sharp.

\begin{table}[!htb]
\caption{State-of-the-art performance comparison on the NYU
dataset}
\huge
\resizebox{.95\linewidth}{!}
{
\begin{tabular}{ | l | c | c | c  c  c | c  c  c |}
\hline
\multirow{2}{*}{{{method}}}                &\multirow{2}{*}{{{\# train}}} &\multirow{2}{*}{{{\# params}}} &\multicolumn{3}{c|}{Accuracy (\%)} &\multicolumn{3}{c|}{Error}  \\
\cline{4-9}
                                           &{} &{} &$\delta < 1.25$ &$\delta < 1.25^2$ &$\delta < 1.25^3$ &Rel &log10 &Rms\\
\hline
Li \etal~\cite{li2015depth}                           &{795}     &{60M} &{62.07}   &{88.61}   &{96.78}   &{0.232}    &0.094      &0.821\\
Liu \etal~\cite{liu2015deep}                          &{795}     &{130M} &{65.0}    &{90.6}    &{97.6}    &{0.213}    &0.087      &0.759\\
Wang \etal~\cite{wang2015towards}                      &{795}     &{-} &{60.5} &{89.0} &{97.0}
&{0.220} &{0.094} &{0.745}\\
Eigen \etal~\cite{eigen2015predicting}          &{240k} &{200M} &76.9      &95.0      &98.8      &0.158      &{-}        &0.641\\
Cao \etal~\cite{cao2016estimating}           &{240k} &{350M} &80.0      &95.6      &98.8      &0.148      &0.063 &{0.615}\\
\hline
ours                                     &\bf{12k}  &{60M} &\bf{83.5} &\bf{96.7} &\bf{99.1} &\bf{0.125} &\bf{0.052} &\bf{0.519}\\
\hline
\end{tabular}}
\label{tab:nyu2}
\end{table}

\begin{table}[!htb]
\huge
\caption{Component evaluation with different settings}
\resizebox{0.8\linewidth}{!}
{
\begin{tabular}{ | l | c  c  c | c  c  c |}
\hline
\multirow{2}{*}{{{method}}}                &\multicolumn{3}{c|}{Accuracy (\%)} &\multicolumn{3}{c|}{Error}  \\
\cline{2-7}
                                           &$\delta < 1.25$ &$\delta < 1.25^2$ &$\delta < 1.25^3$ &Rel &log10 &Rms\\
\hline
no dilation                &{79.9} &{95.8} &{98.5} &{0.139} &{0.061} &{0.589}\\
no skip connection        &{82.1} &{95.8} &{98.8} &{0.132} &{0.055} &{0.523}\\
hard threshold                           &{83.2} &{96.5} &{98.9} &{0.134} &{0.056} &{0.556}\\
\hline
ours                                &\bf{83.5} &\bf{96.7} &\bf{99.1} &\bf{0.125} &\bf{0.052} &\bf{0.519}\\
\hline
\end{tabular}}
\label{tab:analysis}
\end{table}
\section{Conclusion}
An adapted deep convolutional neural network architecture is proposed for single image depth estimation. We also propose the soft-weight-sum inference instead of the hard-threshold method. Experimental results demonstrate that our proposed method achieves better performance than other state-of-the-art methods on NYU Depth V2 dataset.


\vskip3pt

{
\bibliographystyle{IEEEbib}
\bibliography{mybibfile}

\begin{thebibliography}{1}

\bibitem{li2015depth}
Bo~Li, Chunhua Shen, Yuchao Dai, A.~van~den Hengel, and Mingyi He,
\newblock ``Depth and surface normal estimation from monocular images using
  regression on deep features and hierarchical crfs,''
\newblock in {\em CVPR}, jun 2015, pp. 1119--1127.

\bibitem{liu2015deep}
Fayao Liu, Chunhua Shen, Guosheng Lin, and Ian Reid,
\newblock ``Learning depth from single monocular images using deep
  convolutional neural fields,''
\newblock {\em TPAMI}, vol. 38, no. 10, pp. 2024--2039, 2016.

\bibitem{wang2015towards}
Peng Wang, Xiaohui Shen, Zhe Lin, Scott Cohen, Brian Price, and Alan~L Yuille,
\newblock ``Towards unified depth and semantic prediction from a single
  image,''
\newblock in {\em CVPR}, 2015, pp. 2800--2809.

\bibitem{eigen2015predicting}
David Eigen and Rob Fergus,
\newblock ``Predicting depth, surface normals and semantic labels with a common
  multi-scale convolutional architecture,''
\newblock in {\em ICCV}, 2015, pp. 2650--2658.

\bibitem{cao2016estimating}
Yuanzhouhan Cao, Zifeng Wu, and Chunhua Shen,
\newblock ``Estimating depth from monocular images as classification using deep
  fully convolutional residual networks,''
\newblock {\em [Online]. Avaliable: https://arxiv.org/abs/1605.02305}, 2016.

\bibitem{Silberman2012Indoor}
Nathan Silberman, Derek Hoiem, Pushmeet Kohli, and Rob Fergus,
\newblock ``Indoor segmentation and support inference from rgbd images,''
\newblock in {\em ECCV}, 2012, pp. 746--760.

\bibitem{ResNet}
Kaiming He, Xiangyu Zhang, Shaoqing Ren, and Jian Sun,
\newblock ``Deep residual learning for image recognition,''
\newblock in {\em CVPR}, 2016, pp. 770--778.

\bibitem{yu2015multi}
Fisher Yu and Vladlen Koltun,
\newblock ``Multi-scale context aggregation by dilated convolutions,''
\newblock in {\em ICLR}, 2016, pp. 1--10.

\bibitem{jia2014caffe}
Yangqing Jia, Evan Shelhamer, Jeff Donahue, Sergey Karayev, Jonathan Long, Ross
  Girshick, Sergio Guadarrama, and Trevor Darrell,
\newblock ``Caffe: Convolutional architecture for fast feature embedding,''
\newblock in {\em Proc. {ACM} Int. Conf. Multimedia}, 2014, pp. 675--678.

\end{thebibliography}
}
\end{document}